\documentclass[sigconf]{acmart}

\copyrightyear{2021}
\acmYear{2021}
\setcopyright{acmlicensed}\acmConference[CIKM '21]{Proceedings of the 30th
ACM International Conference on Information and Knowledge
Management}{November 1--5, 2021}{Virtual Event, QLD, Australia}
\acmBooktitle{Proceedings of the 30th ACM International Conference on
Information and Knowledge Management (CIKM '21), November 1--5, 2021,
Virtual Event, QLD, Australia}
\acmPrice{15.00}
\acmDOI{10.1145/3459637.3482125}
\acmISBN{978-1-4503-8446-9/21/11}

\RequirePackage{natbib}
\usepackage{enumitem}
\RequirePackage[skip=.8ex plus.2ex]{caption}

\RequirePackage{tikz}
\RequirePackage{pgfplots}
\pgfplotsset{compat=newest}
\usepgfplotslibrary{groupplots}
\usepgfplotslibrary{fillbetween}

\RequirePackage[color=yellow!10,textsize=scriptsize,linecolor=gray!80]{todonotes} % ,disable
\usepackage{amsmath}
\usepackage{multirow}
\usepackage{bm}
\usepackage{numprint}
\RequirePackage{Definitions}

\usepackage{algorithm2e}

\makeatletter
\g@addto@macro{\normalsize}{%
\setlength{\abovedisplayskip}{0pt plus2pt}%
\setlength{\abovedisplayshortskip}{0pt plus2pt}%
\setlength{\belowdisplayskip}{0pt plus2pt}%
\setlength{\belowdisplayshortskip}{0pt plus2pt}}
\makeatother

\RequirePackage{balance}

\newcommand{\set}[1]{\left\{#1 \right\}}
\newcommand{\given}{\,|\,}
\newcommand{\nbr}{\text{nbr}}
\newcommand{\nnbr}{\overline{\text{nbr}}}
\newcommand{\viz}{\emph{viz.}}
\newcommand{\twitterf}{Twitter-1}
\newcommand{\twitters}{Twitter-2}
\newcommand{\twittert}{Twitter-3}

\newcommand{\google}{Google$+$}
\newcommand{\emb}{\texttt{emb}}
\graphicspath{{./Figures/}}
\usepackage{stackengine}
\usepackage{subfig}

\usepackage{float}
\begin{document}
\fancyhead{}

\title{Integrating Transductive And Inductive Embeddings\\ Improves Link Prediction Accuracy}

%% AUTHORS
\author{Chitrank Gupta}
\authornote{Both authors contributed equally to this research.}

\author{Yash Jain}
\authornotemark[1]

\author{Abir De}
\authornote{Contact author \path{abir@cse.iitb.ac.in}}

\author{Soumen Chakrabarti}
\affiliation{%
  \institution{IIT Bombay}
  \country{India}
}

% \author[1]{Chitrank Gupta}
% \author[1]{Yash Jain}
% \author[2]{Abir De}
% \authornote{sdf}
% \author[3]{Author C}
% % \author[2]{Author D}
% % \author[2]{Author E}
% \affiliation[ ]{Department of Computer Science,  University}
% % \affil{Department of Mechanical Engineering, University}
% \affiliation[ ]{\textit {\{email1,email2,email3,email4,email5\}@xyz.edu}}
% % \setcounter{Maxaffil}{0}
% % \renewcommand\Affilfont{\itshape\small}

\begin{abstract}
In recent years, inductive graph embedding models, \emph{viz.}, graph neural networks (GNNs) have become increasingly accurate at link prediction (LP) in online social networks. The performance of such networks depends strongly on the input node features, which vary across networks and applications. 
Selecting appropriate node features remains application-dependent and generally an open question. Moreover, owing to privacy and ethical issues, use of personalized node features is often restricted. In fact, many publicly available data from online social network do not contain any node features (e.g., demography). In this work, we provide a comprehensive experimental analysis which shows that harnessing a transductive technique (e.g., Node2Vec) for obtaining initial node representations, after which an inductive node embedding technique takes over, leads to substantial improvements in link prediction accuracy. We demonstrate that, for a wide variety of GNN variants, node representation vectors obtained from Node2Vec serve as high quality input features to GNNs, thereby improving LP performance.
\end{abstract}
% \keywords{transductive model; inductive model; link prediction}
\maketitle

\section{Introduction}
\label{sec:introduction}

    The link prediction problem (LP) is to predict the set of future edges in a graph which are more likely to appear than the other edges, after observing the graph at a certain time. It has a wide variety of applications, \viz, suggesting friends in social networks, connecting people on LinkedIn~\cite{nowell-kleinberg}, recommending movies 
    on Netflix~\cite{chen2005link,li2009recommendation}, etc.  LP has been thoroughly studied from the perspectives of both network science and machine learning~\cite{Katz1953,Katz1997start, kipf2016semi,hamilton2017inductive, xu2018powerful, ZhangC2018LinkPredGNN, roy2021adversarial} starting from the seminal papers by~\citet{Lada} and~\citet{nowell-kleinberg}.

\subsection{Prior work and limitations}

In recent years, there has been a flurry of deep learning models which have shown a great potential in predicting links in online social networks. Such models learn node embedding vectors which compress sparse, high dimensional information about node neighborhoods in the graph into low dimensional dense vectors. These embedding methods predominantly follow two approaches. The first approach consists of transductive neural models which learn the embedding of each node separately by leveraging matrix factorization based models, \eg, Node2Vec~\cite{grover2016node2vec}, DeepWalk~\cite{perozzi2014deepwalk}, etc. Such models suffer from two major limitations, \viz, they cannot make predictions using unseen nodes and, the corresponding model complexity grows with 
the network size~\cite{hamilton2017inductive}. On the other hand, the second approach consists of graph neural networks (GNNs) which train  
inductive neural models using symmetric aggregator functions~\cite{kipf2016semi,ravanbakhsh2016deep,zaheer2017deep,wu2019simplifying,schlichtkrull2018rgcn,yadati2018hyperlp,corso2020principal,morris2020weisfeiler}.
% \footnote{Although some proposals like GAT \cite{velivckovic2017graph} and UniMP \cite{shi2021masked} combine the node information asymmetrically using attention weights, the weights themselves are computed via symmetric functions.}.
Such models feed the information about the node-neighborhoods into a symmetric aggregator which shares the same parameters globally across the graph. Hence, they can predict links on unseen nodes which were not present during training.

Given a node $u$, a GNN model collects features of the nodes at different distances from  $u$ and combine them
using a symmetric set aggregator. Therefore, the predictive prowess of GNN is contingent on providing it appropriate node features, which however, are domain dependent and require significant effort to design. Moreover, in the context of link recommendation in online social networks, the use of node features may be restricted due to privacy and ethical issues. As a result, most publicly available datasets on online social networks severely limit access to node features. Therefore, existing works use different proxies, \eg, random features, one-hot encoding of node-IDs, etc.

% \vspace{-3mm}
\subsection{Our work}
% \vspace{-1mm}
In this work, we provide a detailed experimental study where
we first train one set of node embeddings which are agnostic to the graph neural network (GNN), using a transductive embedding model. Such node embeddings contain only structural information about a node. Next, we feed these embedding vectors as input node features into an inductive model. Finally, we train this inductive model using an LP loss function, to obtain the final node embeddings. We observed that this two step procedure is consistently more effective than introducing other types of ad-hoc proxy node features into the model. 

We perform a comparative analysis across seven datasets and four GNN models, which reveal that for online social networks, such a combination of transductive and inductive models consistently improves predictive performance, for all commonly-used LP loss functions\footnote{Our code is available at \url{https://www.cse.iitb.ac.in/~abir/codes/linkPredCIKM2021.zip}}. However, for citation graphs, such an approach does not provide any improvement, which indicates that the structural properties alone are not enough to capture the generative process of the underlying graph. 
% We release our code  here\footnote{\url{https://www.cse.iitb.ac.in/~abir/codes/linkPredCIKM2021.zip}}.

% \href{https://github.com/theartpiece/Link_Prediction_Study/}{git repository}\footnote{\url{https://github.com/theartpiece/Link_Prediction_Study/}}.

\section{Link Prediction Methods}
\label{sec:Methods}

After setting up notation and preliminaries, we will introduce the two major mechanisms through which node representations can be obtained: inductive and transductive embeddings.  Then we will describe how to combine their strengths.   We will conclude the section with a brief review on standard loss objectives for LP.

\subsection{Notation}

Given an undirected graph $G=(V,E)$, we denote $\nbr(u)$ and $\nnbr(u)$ as the neighbors and non-neighbors of $u$, respectively. In this context, note that $\nbr(u)$ also contains $u$, \ie, $\nbr(u)=\set{v\given (u,v)\in E}\cup \set{u}$, $\nnbr(u)=\set{v\neq u \given (u,v)\not\in E}$. Finally, we denote $y_{u,v}=1$ if $v\in \nbr(u)$ and $y_{u,v} = 0$, otherwise.

\subsection{Inductive node embeddings}

Suppose, through some means, we obtain an initial node representation $\zb_u$ for each node~$u$.  These can be used by an inductive model to obtain graph-context sensitive node representations.  Various graph neural networks (GNNs) are standard examples of inductive models.
At a very high level, given a node $u$ and an integer $K$,  a GNN takes the underlying graph $G=(V,E)$ and the initial node embeddings $\set{\zb_u \given u \in V }$ as input, and then computes inductive node embeddings $\set{\emb_u \given u \in V }$ by iteratively aggregating structural information from the nodes $k=1,2,...,K$ hops away from~$u$.
\begin{align}
 &\xb_u(0) = \zb_u  \label{eq:gnn-1}\\
 & \xb_u(k) = F_{\theta}(\xb_v(k-1)\given v\in \nbr(u))\label{eq:gnn-2}\\
 & \emb_u = G_{\theta} (\xb_u(k) \given k\in[K]) \label{eq:gnn-3}
\end{align}
In Eqs.~\eqref{eq:gnn-2} \& ~\eqref{eq:gnn-3}, $\xb_u(\bullet)$ are intermediate node embeddings. Moreover, $F_{\theta}$ and $G_{\theta}$ are neural networks parameterized with $\theta$. 
% In particular $F_{\theta}$ is a symmetric aggregator function which ensures that the final embedding remains invariant to the permutation of the nodes. 
The exact form of $F_{\theta}$ and $G_{\theta}$ varies across different GNN models.  In this paper, we consider four competitive GNNs, \viz, GCN~\cite{kipf2016semi}, GraphSAGE~\cite{hamilton2017inductive}, GIN~\cite{xu2018powerful} and DGCNN~\cite{ZhangC2018LinkPredGNN}.
For a downstream task such as LP, we compute the score for the node-pair $(u,v)$~as 
\begin{align}
s_{\theta}(u,v) = H_{\theta}(\emb_u,\emb_v) \label{eq:score}
\end{align}
Here $H_{\theta}$ is another network that compares embeddings $\emb_u, \emb_v$ to arrive at a link score.  We call this scheme `inductive' because, once $F_\theta, G_\theta, H_\theta$ are trained for a task, the induced model can be applied to a completely new graph in the same problem domain, to get meaningful node representations.

\subsection{Transductive node embeddings}

For each application, a key question is how to set initial node representations~$\set{\zb_u}$.  If rich local signals (such as text or demographics) are available at each node, these can be suitably featurized using classical feature engineering or deep techniques.  In the absence of node features, GNN practitioners have tried fixed random vectors, one-hot codes for (arbitrary) node IDs, and a row or column of the adjacency matrix corresponding to each node.  One may argue that the GNNs are in charge of network signal aggregation, so local features should suffice for setting up~$\set{\zb_u}$.  Our key noteworthy observation is that when graph features are used to \emph{also} determine the initial $\set{\zb_u}$, the GNNs may behave better and lead to better end-task accuracy. We will now describe two ways in which how graph-based transductive methods can be used to obtain these initial node representations.

\xhdr{Node2Vec (N2V)~\cite{grover2016node2vec}} Given a node $u$, Node2Vec first samples nodes 
$N_S(u)$ using different random walk based heuristics. Then it models the likelihood of the sampled
nodes using a multinomial distribution informed by the proximity of the nodes, measured in terms of  $\exp(\zb_u ^\top \zb_v)$.
More specifically, we have:
\begin{align}
 \text{Pr} (N_S (u)\given \zb_u ) = \prod_{v\in N_S(u)} \frac{\exp(\zb_u ^\top \zb_v)}{\sum_{w\in V}\exp(\zb_u ^\top \zb_w)}
\end{align}
Finally, the node embeddings $\zb_u$ are estimated by solving the following training problem.
\begin{align}
 \max_{\set{\zb_u}} \sum_{u\in V} \left[ - \log\sum_{w\in V}\exp(\zb_u ^\top \zb_w) +  \sum_{v\in N_S(u)}\zb_u ^\top \zb_v \right]\label{eq:node2vec}
\end{align}

\xhdr{Matrix factorization (MF)~\cite{menon2011matfactor}} Apart from Node2Vec, we also consider
Matrix factorization as a candidate transductive model, \ie, we train the node embeddings
$\set{\zb_u}$ as,
\begin{align}
 \min_{\set{\zb_u}}  \sum_{u\in V}\sum_{v\in V}(y_{u,v}-\zb_u ^\top \zb_v)^2  + \lambda \sum_{u\in V}||\zb_u||^2 _2\label{eq:mf}
\end{align}
where $\lambda$ is a regularizing constant.

\subsection{Combining inductive and transductive approaches}
Once we train the transductive embeddings $\set{\zb_u}$ using Eq.~\eqref{eq:node2vec} or ~\eqref{eq:mf}, we feed them into the GNN model~\eqref{eq:gnn-1}--~\eqref{eq:gnn-3} to learn $\emb_u$.
% \xhdr{Node Features}
% We keep three kinds of node features
% \begin{itemize}
%     \item[1] Node2vec features. These are the unsupervised node features obtained from the node2vec algorithm.
%     \item[2] Attribute features. These are the node features that come along with the dataset. In many of our datasets, becuase we did not have any such node attributes, we simply used the one-hot encoding of the nodes.
%     \item[3] Node labels. These nodes are obtained from the DRNL (ref- SEAL paper) algorithm.
% \end{itemize}

% \xhdr{Feeding input data.}
% We observe that different models in their respective works fed different type of information and different kind of information. So to have a fair comparison among models, we train and test models on three different ways of feeding the data. They are:

% \subsection{Training loss objectives}

\xhdr{Training losses} We learn the parameters $\theta$ used to parameterize Eqs.~\eqref{eq:gnn-2},~\eqref{eq:gnn-3}
by minimizing a LP loss function --- the two most popular of which binary cross entropy (BCE) loss and ranking loss are considered in this work.
% \begin{table}[]
% \begin{tabularx}{|c|c|}
%         & Equation\\
% BCE     & \parbox{3cm}{
% \begin{align}
%  \ell_{\text{BCE}}(\theta; G) &=   
%     -\sum_{(u,v) \in E} \log[ \sigma(s_\theta {(u,v)})]\nn\\
%     &\qquad-\sum_{(u,v)\not\in E}\log[1- \sigma(s_\theta {(u,v)})]
% \end{align}

%                     }\\
% Ranking & \parbox{3cm}{
%     \begin{align}
%  \ell_{\text{Rank}}(\theta; G) \\ =   
%     \sum_{u\in V}\sum_{\substack{w\not\in\nbr(u) \\ v\in \nbr(u)}} \text{ReLU}(s_{\theta}(w,u)-s_{\theta}(v,u)+\delta)
% \end{align}
%                     }
% \end{tabularx}
% \end{table}
\xhdr{BCE loss} We compute the binary cross entropy loss as follows: 
\begin{align}
 \ell_{\text{BCE}}(\theta; G) &=   
    -\sum_{(u,v) \in E} \log[ \sigma(s_\theta {(u,v)})]-\sum_{(u,v)\not\in E}\log[1-\sigma(s_\theta {(u,v)})]
\end{align}

\xhdr{Pairwise ranking loss} We compute the pairwise ranking loss as follows:
\begin{align}
 \ell_{\text{Rank}}(\theta; G) &=   
    \sum_{u\in V}\sum_{\substack{w\not\in\nbr(u) \\ v\in \nbr(u)}} \text{ReLU}(s_{\theta}(w,u)-s_{\theta}(v,u)+\delta)
\end{align}
where $\sigma$ is the \emph{sigmoid} function and
$\delta$ is a tunable margin.

% \begingroup \color{blue}
% \todo{check and discard}
% Loss functions- We train our models using both Binary-cross entropy (BCE) Loss and Ranking loss. Both of these losses operate on a non-normalised score $s$ output for each of the candidate edges.  Because of the difference in how we sample  a batch for training and hence its inevitable effect on the performance, we keep experiments with different loss functions out of competition.
% \endgroup

\begin{table}[th]
\centering 
\resizebox{.8\hsize}{!}{
\begin{tabular}{|l||c|c|c|c|c|c|} \hline
Dataset &$|V|$&$|E|$& $d_{avg}$ & Diameter &  $|\Qcal|$  \\ \hline \hline
\twitterf   & 213  & 12173  & 115.38 & 3    & 209 \\ \hline
\twitters   & 235  & 10862  & 92.44 & 3    & 235 \\ \hline
\twittert   & 193  & 7790  & 79.73 & 4    & 190 \\ \hline
\google   & 769  & 22515  & 57.56  & 7    & 718 \\ \hline
PB   & 1222  & 17936  & 28.36  & 8    & 999 \\ \hline
Citeseer   & 3312  & 7848  & 3.74  & 28    & 1010 \\ \hline
Cora   & 2708  & 7986  & 4.90  & 19    & 1470 \\ \hline
\end{tabular} }
\caption{Dataset statistics.}
\label{tab:datasets}
\end{table}

\section{Experiments}
\label{sec:experiments}
In this section, we provide a comprehensive evaluation of our proposed approach, and comparisons with standard alternatives.

\subsection{Datasets}
\label{sec:datasets}

We use seven datasets from diverse domains. 
Among them, the first three datasets, \viz, \twitterf, \twitters, \twittert\ are three separate connected components from Twitter~\cite{leskovec2012learning}. The other datasets are Google+~\cite{leskovec2010kronecker}, PB~\cite{ackland2005mapping}, Cora~\cite{getoor2005link, sen2008collective} and Citeseer~\cite{getoor2005link, sen2008collective}. Among them, the first five datasets are  online social networks, whereas, the last two datasets are citation networks. Refer Table~\ref{tab:datasets} for details.

\subsection{Setup}
\label{sec:setup}
\xhdr{Candidates for inductive and transductive models}
We consider four candidates for the inductive GNNs --- GCN~\cite{kipf2016semi}, GIN~\cite{xu2018powerful}, DGCNN~\cite{ZhangC2018LinkPredGNN} and GraphSAGE~\cite{hamilton2017inductive}.
On the other hand, we consider two candidates for transductive models--- N2V~\cite{grover2016node2vec} and MF~\cite{menon2011matfactor}.

\xhdr{LP methods}
As described in Section~\ref{sec:Methods},
we first train the transductive node embeddings $\set{\zb_u}$ and then feed them
into the inductive models as input. While doing so, we also augment these transductive embeddings with the one-hot encodings of node labels obtained using Double-Radius Node Labeling (DRNL) algorithm~\cite{ZhangC2018LinkPredGNN,zhang2020revisiting}.
In addition, we also compare our approach with the corresponding inductive model which only uses the DRNL node features.

\xhdr{Implementation details}
We used Adam optimizer with learning rate $10^{-3}$.
We use early stopping during training with patience parameter $P=6$, i.e., we stop training when validation fold performance does not improve within the last $P$ epochs. For the ranking losses, we cross validate our method across three values of the tunable margin $\delta\in\set{0.1,1,10}$. 

\begin{table*}[t]
\centering
\resizebox{\hsize}{!}{%
\tabcolsep 2pt
\begin{tabular}{c|cc|cc|cc|cc||cc|cc|cc|cc}
\hline
 \emph{BCE} & \multicolumn{8}{c||}{\textbf{Mean Average Precision (MAP)}} & \multicolumn{8}{c}{\textbf{Mean Reciprocal Rank (MRR)}} \\ \hline
  Models $\to$ & GCN &  & GIN &  & DGCNN &  & GraphSAGE &  & GCN &  & GIN &  & DGCNN &  & GraphSAGE &  \\ \hline
 & W/o & With & W/o & With & W/o & With & W/o & With & W/o & With & W/o & With & W/o & With & W/o & With \\
  Datasets $\downarrow$  & N2V & N2V & N2V & N2V & N2V & N2V & N2V &
 N2V & N2V & N2V & N2V & N2V & N2V & N2V & N2V & N2V\\ \hline \hline

 \twitterf & 0.845 & \multicolumn{1}{c|}{\textbf{0.878}} & 0.845 & \multicolumn{1}{c|}{\textbf{0.859}} & 0.833 & \multicolumn{1}{c|}{\textbf{0.859}} & 0.847 & \textbf{0.876} & 0.953 & \multicolumn{1}{c|}{\textbf{0.962}} & 0.935 & \multicolumn{1}{c|}{\textbf{0.95}} & 0.939 & \multicolumn{1}{c|}{\textbf{0.952}} & 0.932 & \textbf{0.971} \\ 
\twitters & 0.706 & \multicolumn{1}{c|}{\textbf{0.745}} & 0.712 & \multicolumn{1}{c|}{\textbf{0.744}} & 0.711 & \multicolumn{1}{c|}{\textbf{0.747}} & 0.711 & \textbf{0.754} & 0.926 & \multicolumn{1}{c|}{\textbf{0.945}} & 0.927 & \multicolumn{1}{c|}{\textbf{0.936}} & 0.922 & \multicolumn{1}{c|}{\textbf{0.945}} & 0.919 & \textbf{0.951} \\
\twittert & 0.734 & \multicolumn{1}{c|}{\textbf{0.769}} & 0.743 & \multicolumn{1}{c|}{\textbf{0.773}} & 0.734 & \multicolumn{1}{c|}{\textbf{0.766}} & 0.738 & \textbf{0.781} & 0.878 & \multicolumn{1}{c|}{\textbf{0.909}} & 0.897 & \multicolumn{1}{c|}{\textbf{0.908}} & 0.885 & \multicolumn{1}{c|}{\textbf{0.912}} & 0.878 & \textbf{0.934} \\
PB & 0.436 & \multicolumn{1}{c|}{\textbf{0.445}} & 0.43 & \multicolumn{1}{c|}{\textbf{0.456}} & 0.418 & \multicolumn{1}{c|}{\textbf{0.434}} & 0.425 & \textbf{0.448} & 0.669 & \multicolumn{1}{c|}{\textbf{0.673}} & 0.654 & \multicolumn{1}{c|}{\textbf{0.682}} & 0.637 & \multicolumn{1}{c|}{\textbf{0.663}} & 0.649 & \textbf{0.679} \\
\google\ & 0.585 & \multicolumn{1}{c|}{\textbf{0.615}} & 0.528 & \multicolumn{1}{c|}{\textbf{0.624}} & 0.563 & \multicolumn{1}{c|}{\textbf{0.597}} & 0.364 & \textbf{0.633} & 0.808 & \multicolumn{1}{c|}{\textbf{0.816}} & 0.764 & \multicolumn{1}{c|}{\textbf{0.82}} & 0.794 & \multicolumn{1}{c|}{\textbf{0.797}} & 0.59 & \textbf{0.822} \\
Cora & \textbf{0.5} & \multicolumn{1}{c|}{0.45} & \textbf{0.493} & \multicolumn{1}{c|}{0.465} & \textbf{0.481} & \multicolumn{1}{c|}{0.473} & 0.429 & \textbf{0.495} & \textbf{0.561} & \multicolumn{1}{c|}{0.503} & \textbf{0.554} & \multicolumn{1}{c|}{0.522} & \textbf{0.54} & \multicolumn{1}{c|}{0.518} & 0.483 & \textbf{0.552} \\
Citeseer & \textbf{0.518} & \multicolumn{1}{c|}{0.476} & \textbf{0.521} & \multicolumn{1}{c|}{0.516} & \textbf{0.508} & \multicolumn{1}{c|}{0.469} & 0.48 & \textbf{0.539} & \textbf{0.578} & \multicolumn{1}{c|}{0.524} & \textbf{0.581} & \multicolumn{1}{c|}{0.567} & \textbf{0.572} & \multicolumn{1}{c|}{0.53} & 0.538 & \textbf{0.601} \\

\hline
\end{tabular}%
}
\resizebox{\hsize}{!}{%
\tabcolsep 2pt
\begin{tabular}{c|cc|cc|cc|cc||cc|cc|cc|cc}
\hline
 \emph{Ranking} & \multicolumn{8}{c||}{\textbf{Mean Average Precision (MAP)}} & \multicolumn{8}{c}{\textbf{Mean Reciprocal Rank (MRR)}} \\ \hline
  Models $\to$ & GCN &  & GIN &  & DGCNN &  & GraphSAGE &  & GCN &  & GIN &  & DGCNN &  & GraphSAGE &  \\ \hline
 & W/o & With & W/o & With & W/o & With & W/o & With & W/o & With & W/o & With & W/o & With & W/o & With \\
  Datasets $\downarrow$  & N2V & N2V & N2V & N2V & N2V & N2V & N2V &
 N2V & N2V & N2V & N2V & N2V & N2V & N2V & N2V & N2V\\ \hline \hline

\twitterf & 0.814 & \multicolumn{1}{c|}{\textbf{0.858}} & 0.704 & \multicolumn{1}{c|}{\textbf{0.8}} & 0.832 & \multicolumn{1}{c|}{\textbf{0.84}} & 0.792 & \textbf{0.846} & 0.933 & \multicolumn{1}{c|}{\textbf{0.942}} & 0.776 & \multicolumn{1}{c|}{\textbf{0.898}} & \textbf{0.936} & \multicolumn{1}{c|}{0.932} & 0.868 & \textbf{0.954} \\ 
\twitters & 0.694 & \multicolumn{1}{c|}{\textbf{0.734}} & 0.57 & \multicolumn{1}{c|}{\textbf{0.644}} & 0.703 & \multicolumn{1}{c|}{\textbf{0.781}} & 0.693 & \textbf{0.738} & 0.923 & \multicolumn{1}{c|}{\textbf{0.928}} & 0.761 & \multicolumn{1}{c|}{\textbf{0.8}} & 0.92 & \multicolumn{1}{c|}{\textbf{0.922}} & 0.897 & \textbf{0.928} \\
\twittert & 0.732 & \multicolumn{1}{c|}{\textbf{0.771}} & 0.603 & \multicolumn{1}{c|}{\textbf{0.738}} & 0.741 & \multicolumn{1}{c|}{\textbf{0.768}} & 0.737 & \textbf{0.767} & 0.882 & \multicolumn{1}{c|}{\textbf{0.907}} & 0.777 & \multicolumn{1}{c|}{\textbf{0.888}} & 0.896 & \multicolumn{1}{c|}{\textbf{0.915}} & 0.888 & \textbf{0.914} \\
PB & 0.43 & \multicolumn{1}{c|}{\textbf{0.436}} & 0.222 & \multicolumn{1}{c|}{\textbf{0.36}} & 0.414 & \multicolumn{1}{c|}{\textbf{0.426}} & 0.276 & \textbf{0.449} & 0.662 & \multicolumn{1}{c|}{\textbf{0.662}} & 0.374 & \multicolumn{1}{c|}{\textbf{0.584}} & 0.635 & \multicolumn{1}{c|}{\textbf{0.649}} & 0.451 & \textbf{0.676} \\
\google\ & 0.513 & \multicolumn{1}{c|}{\textbf{0.576}} & 0.262 & \multicolumn{1}{c|}{\textbf{0.34}} & 0.456 & \multicolumn{1}{c|}{\textbf{0.48}} & 0.379 & \textbf{0.611} & 0.763 & \multicolumn{1}{c|}{\textbf{0.777}} & 0.477 & \multicolumn{1}{c|}{\textbf{0.566}} & \textbf{0.703} & \multicolumn{1}{c|}{0.696} & 0.54 & \textbf{0.801} \\
Cora & 0.479 & \multicolumn{1}{c|}{\textbf{0.486}} & 0.424 & \multicolumn{1}{c|}{\textbf{0.427}} & \textbf{0.475} & \multicolumn{1}{c|}{0.437} & 0.384 & \textbf{0.45} & 0.533 & \multicolumn{1}{c|}{\textbf{0.54}} & \textbf{0.482} & \multicolumn{1}{c|}{0.479} & \textbf{0.53} & \multicolumn{1}{c|}{0.478} & 0.434 & \textbf{0.506} \\
Citeseer & \textbf{0.516} & \multicolumn{1}{c|}{0.483} & 0.429 & \multicolumn{1}{c|}{\textbf{0.468}} & \textbf{0.498} & \multicolumn{1}{c|}{0.484} & \textbf{0.49} & 0.479 & \textbf{0.584} & \multicolumn{1}{c|}{0.537} & 0.474 & \multicolumn{1}{c|}{\textbf{0.517}} & \textbf{0.562} & \multicolumn{1}{c|}{0.548} & \textbf{0.546} & 0.541 \\

\hline
\end{tabular}%
}
\caption{Performance comparison in terms of MAP (left half) and the MRR (right half) of our proposed method --- which integrates the transductive and inductive models --- against the inductive model which does not receive any input from the transductive model, across all datasets and all inductive models, \ie, GCN~\cite{kipf2016semi}, GIN~\cite{xu2018powerful}, DGCNN~\cite{ZhangC2018LinkPredGNN} and GraphSAGE~\cite{hamilton2017inductive}. We choose N2V as the transductive embedding model. The top and the bottom halves of the table report the results for BCE loss and Ranking loss objective, respectively. We observe that for all datasets except citation networks, our proposed approach outperforms the inductive model trained without transductive embeddings.}
\label{tab:main}
\end{table*}

\subsection{Evaluation}

\xhdr{Protocol}
\label{sec:protocol}
As suggested in previous works~\cite{roy2021adversarial,BackstromL2011SRW}, we consider predicting 
only on those node pairs whose one of the nodes participate in at least one triangle. We call such nodes query nodes $Q$. Then for each query node $q\in Q$, we randomly
split both $\nbr(q)$ and $\nnbr(q)$ into 70\% training, 10\% validation and 20\% test sets. We use the training set to supervise training of the GNN model. Next, we rank the node pairs in the test set based on the 
scores computed by the trained LP model. 

\xhdr{Metrics}
We evaluate the predicted ranked list (of node pairs belonging to a query node $q \in Q$)
via average precision (AP) and reciprocal rank (RR).
Finally, we report Mean Average Precision (MAP) and Mean Reciprocal Rank (MRR) as follows:
\begin{align}
    \text{MAP}=\frac{1}{|Q|}\sum_{q\in Q} AP_q, \quad 
    \text{MRR}=\frac{1}{|Q|}\sum_{q\in Q} RR_q,
\end{align}
where. $AP_q$ and $RR_q$ are the average precision and reciprocal rank corresponding to the ranked list given by $q\in Q$.
%

% We evaluate all our models on per-query averaging metric. For that, we first sample a subset of nodes as query nodes. Next for each of these query nodes we select only a subset of non-neighboring nodes as the candidate future nodes. The qualification of a query node is that it must participate in at least one triangle. We independently partition the neighboring or non-neighboring neighbors to obtain the training, validation and test splits for every query node. Later on we redistribute them to maintain the constraint that no edge or non-edge appears in the two different splits. The conscience behind choosing only 2-hop neighbors as future candidate neighbors (we have to cite this right?) is the reason that a huge number of new edges generally close a triad with edges already already formed. Such properties may break after redistributing the edges in various splits, but this only makes the nodes of such non-edges go further apart. 

% \subsubsection{Metrics}
% \label{sec:metrics}

% We observe that for every query node there would be (and very rarely otherwise) many true neighbors. Hence metrics like Hits@K or MRR are insufficient to truly capture the performance. Instead, we measure the performance via mAP. To calculate mAP, we average over the AP on the ranked list of candidate neighbors for every query node.

\subsection{Results}

\begin{figure}[h]
\centering

\subfloat[\google]{
\includegraphics[width=.21\textwidth]{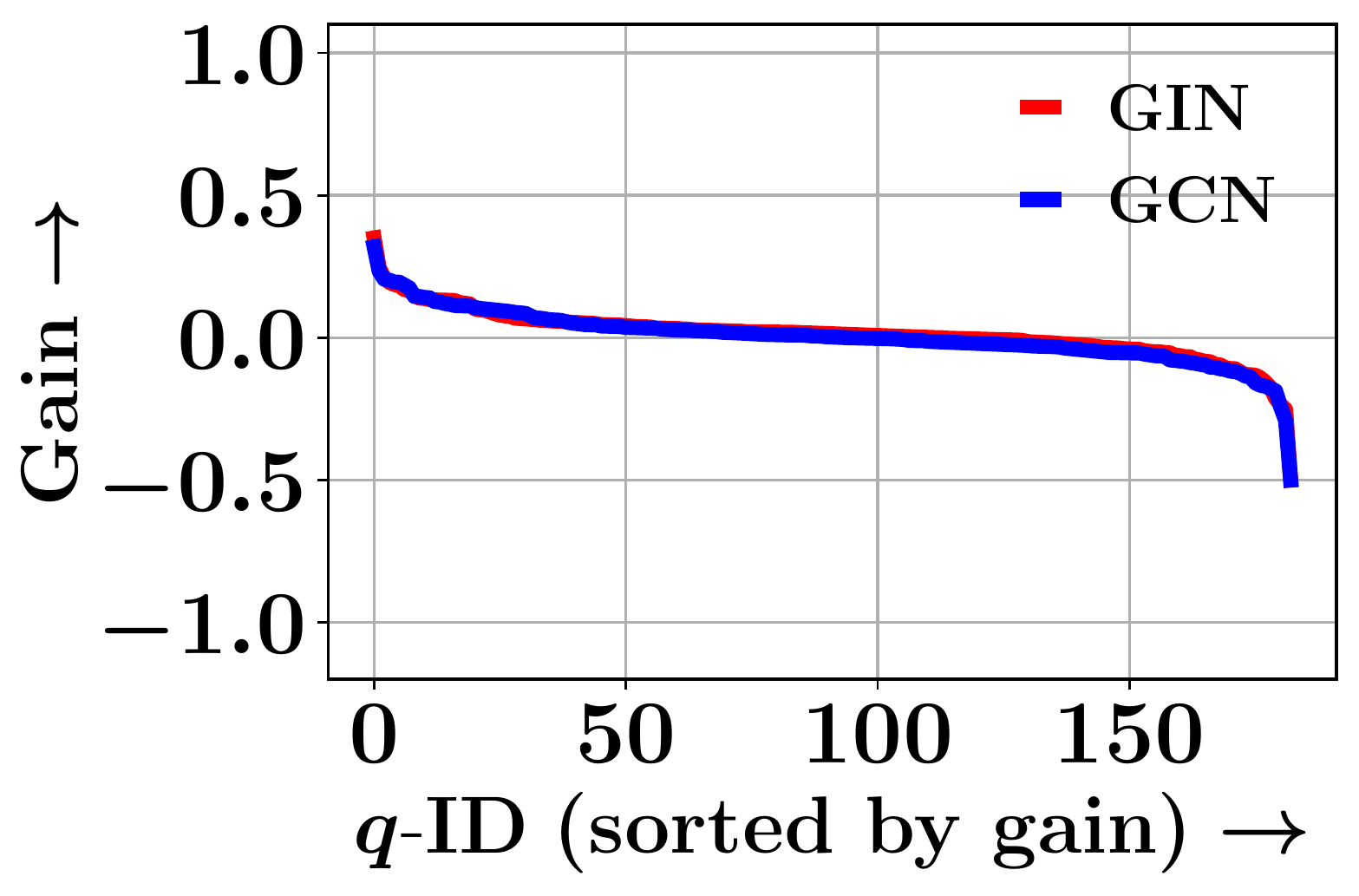}
}\hspace{2mm}
\subfloat[\twittert]{
\includegraphics[width=.21\textwidth]{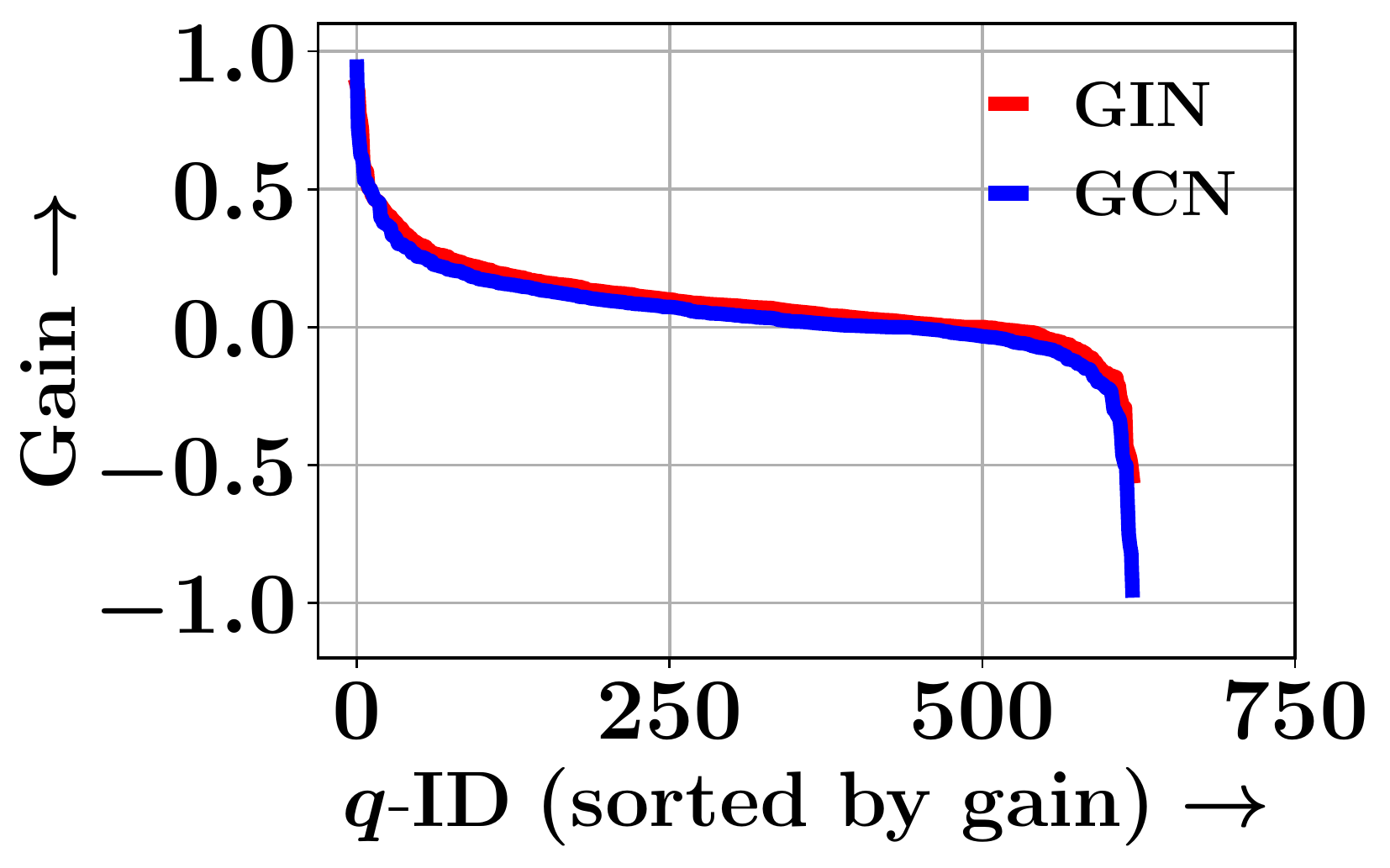}
}
\caption{Query node wise performance comparison in terms of $\text{Gain}=AP(\text{Our method})-AP(\text{GNN-Only})$, \ie, the gain in AP achieved by our method against the corresponding GNN model which does not receive any input from the transductive model. Queries on $x$-axis are sorted in the decreasing order of gain.   }
\label{fig:MapMrrDiff}
\end{figure}

\xhdr{Comparative analysis} First,  we compare our approach against the corresponding inductive embedding model trained only with the DRNL features. Here, we consider N2V as the transductive embedding model. Table~\ref{tab:main} summarizes the results for both BCE loss and Ranking loss, which shows that:
(i)~for all datasets, except for citation graphs, our method outperforms the other method;
(ii)~the trained N2V embeddings provide significant performance boost in GraphSAGE;
and, (iii)~the performance of GCN and GraphSAGE are comparable in the absence of N2V embeddings, as they
share similar neural architectures.
N2V computes the node embeddings by performing long range random walk, whereas, the GNN models limit their aggregation operation within $K\le 3$ hop distance. As a result, our approach is able to capture the structural information better than a GNN model trained alone with the DRNL features. 
\xhdr{Query-wise analysis} Next, we look into the performance at the individual query level.  Specifically, we probe the gain/loss achieved by our model in terms of $\text{Gain}=AP(\text{Our method})-AP(\text{GNN-Only})$ for each query $q\in Q$. Figure~\ref{fig:MapMrrDiff} summarizes the results which show that, for GCN model (GIN model), our method provides superior performance for 71\%  and 53\% (81\% and 61\%) queries for \twittert\ and \google\ datasets, respectively.

\xhdr{Effect of Matrix Factorization (MF) as the inductive model}
Then, we investigate if the superior performance of our approach is consistent across transductive models. We show this to be the case, when MF is the candidate transductive model in Table~\ref{tab:mf}.
\begin{table}[h]
\centering
\resizebox{\hsize}{!}{%
\tabcolsep 2pt
\begin{tabular}{c|cc|cc|cc|cc}
\hline
 Models $\to$ & \multicolumn{2}{c|}{GCN} &  \multicolumn{2}{c|}{GIN}   & \multicolumn{2}{c|}{DGCNN} & \multicolumn{2}{c}{GraphSAGE}  \\ \hline
 & W/o & With & W/o & With & W/o & With & W/o & With  \\
  Datasets $\downarrow$  & MF & MF & MF & MF & MF & MF & MF &
 MF \\ \hline \hline

\twitterf & 0.845 & \multicolumn{1}{r|}{\textbf{0.859}} & 0.845 & \multicolumn{1}{r|}{\textbf{0.851}} & 0.833 & \multicolumn{1}{r|}{\textbf{0.857}} & 0.847 & \textbf{0.872} \\ 
\twitters & 0.706 & \multicolumn{1}{r|}{\textbf{0.734}} & 0.712 & \multicolumn{1}{r|}{\textbf{0.736}} & 0.711 & \multicolumn{1}{r|}{\textbf{0.739}} & 0.711 & \textbf{0.752} \\
\twittert & 0.734 & \multicolumn{1}{r|}{\textbf{0.766}} & 0.743 & \multicolumn{1}{r|}{\textbf{0.775}} & 0.734 & \multicolumn{1}{r|}{\textbf{0.755}} & 0.738 & \textbf{0.781} \\
PB & 0.436 & \multicolumn{1}{r|}{\textbf{0.457}} & 0.43 & \multicolumn{1}{r|}{\textbf{0.452}} & 0.418 & \multicolumn{1}{r|}{\textbf{0.45}} & 0.425 & \textbf{0.438} \\
\google\ & 0.585 & \multicolumn{1}{l|}{\textbf{0.624}} & 0.528 & \multicolumn{1}{r|}{\textbf{0.616}} & 0.563 & \multicolumn{1}{r|}{\textbf{0.585}} & 0.364 & \multicolumn{1}{r}{\textbf{0.637}} \\
\hline
\end{tabular}%
}
    \caption{Performance comparison in terms of MAP of our proposal when we use MF~\cite{menon2011matfactor} as the transductive model against the inductive model trained without any transductive embedding for the first five datasets with BCE loss.}
\label{tab:mf}
\end{table}
\xhdr{Effect of raw node attributes instead of N2V features} 
Recall that, the alternative GNN models use the node features computed via DRNL algorithm~\cite{ZhangC2018LinkPredGNN,zhang2020revisiting}. Here, we augment them with the 
available node attributes and compare its performance with our proposed approach. We find out that the performance of the two methods is comparable, as shown in Table~\ref{tab:raw-feat}.
% 	N2V (OLD) only	N2V (NEW) only
% Twitter_3	0.769	0.775
% Gplus_1	0.615	0.618

\begin{table}[t]
\centering
\resizebox{\hsize}{!}{%
\tabcolsep 2pt
\begin{tabular}{c|cc|cc|cc|cc}
\hline
 & \multicolumn{8}{c}{\textbf{Mean Average Precision (MAP)}} \\ \hline
  Models $\to$ & \multicolumn{2}{c|}{GCN} &  \multicolumn{2}{c|}{GIN}   & \multicolumn{2}{c|}{DGCNN} & \multicolumn{2}{c}{GraphSAGE} \\ \hline
  Datasets $\downarrow$ & Attribute & N2V & Attribute & N2V & Attribute & N2V & Attribute & N2V \\
   \hline \hline

\twitterf & 0.871 & \multicolumn{1}{r|}{0.878} & 0.873 & \multicolumn{1}{r|}{0.859} & 0.866 & \multicolumn{1}{r|}{0.859} & 0.873 & 0.876 \\
\twitters & 0.748 & \multicolumn{1}{r|}{0.745} & 0.756 & \multicolumn{1}{r|}{0.744} & 0.743 & \multicolumn{1}{r|}{0.747} & 0.753 & 0.754 \\
\twittert & 0.78 & \multicolumn{1}{r|}{0.769} & 0.775 & \multicolumn{1}{r|}{0.773} & 0.768 & \multicolumn{1}{r|}{0.766} & 0.77 & 0.781 \\
PB & 0.454 & \multicolumn{1}{r|}{0.445} & 0.461 & \multicolumn{1}{r|}{0.456} & 0.451 & \multicolumn{1}{r|}{0.434} & 0.454 & 0.448 \\
\google\ & 0.641 & \multicolumn{1}{r|}{0.615} & 0.642 & \multicolumn{1}{r|}{0.624} & 0.622 & \multicolumn{1}{r|}{0.597} & 0.629 & 0.633 \\
Cora & 0.491 & \multicolumn{1}{r|}{0.45} & 0.462 & \multicolumn{1}{r|}{0.465} & 0.499 & \multicolumn{1}{r|}{0.473} & 0.542 & 0.495 \\
Citeseer & 0.477 & \multicolumn{1}{r|}{0.476} & 0.495 & \multicolumn{1}{r|}{0.516} & 0.502 & \multicolumn{1}{r|}{0.469} & 0.558 & 0.539 \\

\hline
\end{tabular}%
}
 \caption{Performance in terms of MAP of our proposed method  against the inductive model trained with the raw node attributes and DRNL  as the input features. We use BCE loss as training objective. In almost all cases, the performance of the two methods are comparable.}
\label{tab:raw-feat}
\end{table}

\begin{table}[H]
\centering
%\resizebox{0.3\textwidth}{!}{
\begin{tabular}{c|cc}
\hline
&  $r = 0.02|V|$ & $r=0.05|V|$ \\ \hline
\twittert& 0.769 	&0.775 \\ \hline
\google & 0.615 &	0.618 \\ \hline
\end{tabular}
%}
\caption{Effect of length of random walk $r$ in N2V on MAP for \twittert\ and \google\ datasets with GCN model (our approach) and BCE loss.}\label{tab:rw}
\end{table}

\xhdr{Effect of random walk length in Node2Vec}
Finally, we change the length $r$ of the random walk in N2V and observe the MAP values obtained by our approach. Table~\ref{tab:rw} summarizes the results, which shows that the MAP values improves with increasing the value of $r$. This is because  a larger value of $r$ can  encode the structural information of the graph neighborhood of node $u$ better into the node embedding $\set{\zb_u}$ compared to smaller values of~$r$.

\section{Conclusion}
\label{sec:conclusions}

There are two dominant paradigms to represent graph nodes using continuous embedding vectors.  The transductive approach, typified by Node2Vec and Matrix Factorization, scales up the number of parameters with the number of nodes, but can effectively exploit long-range influence in the graphs.  The inductive approach has a globally-tied, smaller-capacity local neighborhood feature aggregator that is rarely applied beyond two hops.  In this paper we establish that combining their strengths can give notable accuracy improvements for social networks where access to intrinsic node features may be restricted or prohibited.

\xhdr{Acknowledgment:}
Both SC and AD are partly supported by an IBM AI Horizon Network grant.
SC is partly supported by a J.\,C.\,Bose Fellowship. AD is partly supported by
DST Inspire grant.

\balance
\bibliographystyle{ACM-Reference-Format}
\bibliography{refs,voila}

\end{document}